\documentclass[sigconf]{acmart}
\AtBeginDocument{%
  }

\setcopyright{acmlicensed}
\copyrightyear{2026}
\acmYear{2026}
\acmDOI{XXXXXXX.XXXXXXX}
\acmConference[MM '26]{Proceedings of the 34th ACM International Conference on Multimedia}{10–14 November 2026}{Rio de Janeiro, Brazil}
\acmISBN{978-1-4503-XXXX-X/2018/06}

\usepackage{amsthm}

\usepackage{bm}
\usepackage{multirow}

\begin{document}

\title{M$^2$GDT: MLLM-Guided Diffusion Transformer with
 Relation-Adaptive Mixture-of-Experts \\for Multimodal Knowledge Graph Completion}
% \author{Anonymous Submission}
\author{Xu Hou}
\email{houx_upc@163.com}
\orcid{}
\affiliation{%
  \institution{Beijing University of Posts and Telecommunications}
  \city{Beijing}
  \country{China}
}

\author{Meiyu Liang}
\affiliation{%
  \institution{School of Computer Science, Beijing University of Posts and Telecommunications}
  \city{Beijing}
  \country{China}
}

\author{Wei Huang}
\affiliation{%
  \institution{Zhejiang University}
  \city{Hangzhou}
  \country{China}
}
\affiliation{%
  \institution{Beijing University of Posts and Telecommunications}
  \city{Beijing}
  \country{China}
}

\author{Yawen Li}
\affiliation{%
  \institution{Beijing University of Posts and Telecommunications}
  \city{Beijing}
  \country{China}
}

\author{Zhe Xue}
\affiliation{%
  \institution{School of Computer Science, Beijing University of Posts and Telecommunications}
  \city{Beijing}
  \country{China}
}

\author{Wu Liu}
\affiliation{%
  \institution{University of Science and Technology of China}
  \city{Hefei}
  \country{China}
}

\author{Guanhua Ye}
\affiliation{%
  \institution{School of Computer Science (National Pilot Software Engineering School), Beijing University of Posts and Telecommunications}
  \city{Beijing}
  \country{China}
}

\author{Lei Shi}
\affiliation{%
  \institution{Communication University of China}
  \city{Beijing}
  \country{China}
}

\author{Kangkang Lu}
\affiliation{%
  \institution{Beijing University of Posts and Telecommunications}
  \city{Beijing}
  \country{China}
}

% \affiliation{%

\renewcommand{\shortauthors}{Anonymous Submission}

\begin{abstract}
    Multimodal Knowledge Graph Completion (MKGC) requires inferring missing entities from structural, textual, and visual cues. Existing diffusion-based MKGC methods usually denoise directly on raw multimodal features. Such a design forces the denoiser to simultaneously perform relation-dependent cue selection, cross-modal semantic alignment, and structure-aware entity generation, which introduces noisy and semantically inconsistent conditions for diffusion and consequently leads to suboptimal completion performance. To address this limitation, we propose \textbf{M$^2$GDT}: \textbf{M}LLM-\textbf{G}uided \textbf{D}iffusion \textbf{T}ransformer with Relation-Adaptive \textbf{M}ixture-of-Experts (\textbf{M$^2$GDT}), a novel MKGC framework built on an \textit{align-then-diffuse} paradigm. M$^2$GDT first employs a \textbf{R}elation-\textbf{A}daptive \textbf{S}emantic \textbf{R}outing \textbf{M}ixture-\textbf{o}f-\textbf{E}xperts (\textbf{RASR-MoE}) module to select relation-relevant multimodal semantic transformation paths and suppress irrelevant modality interference. M$^2$GDT then uses a frozen Multimodal Large Language Model (MLLM) as a semantic anchor to align the routed multimodal representations into a unified latent space and reduce cross-modal semantic heterogeneity. Finally, a \textbf{K}nowledge \textbf{G}raph \textbf{D}iffusion \textbf{T}ransformer (\textbf{KGDT}) performs graph-conditioned denoising generation in the aligned space to produce the missing entity representation. Experiments on three benchmark datasets show that M$^2$GDT consistently outperforms strong baselines.
\end{abstract}


%    <concept>
%        <concept_id>10010147.10010178.10010187.10010198</concept_id>
%        <concept_desc>Computing methodologies~Reasoning about belief and knowledge</concept_desc>
%        <concept_significance>500</concept_significance>
%        </concept>
%  </ccs2012>

% \ccsdesc[500]{Computing methodologies~Reasoning about belief and knowledge}


\maketitle



\section{Introduction}
\begin{figure}[t]
    \centering
    \includegraphics[width=1.0\linewidth]{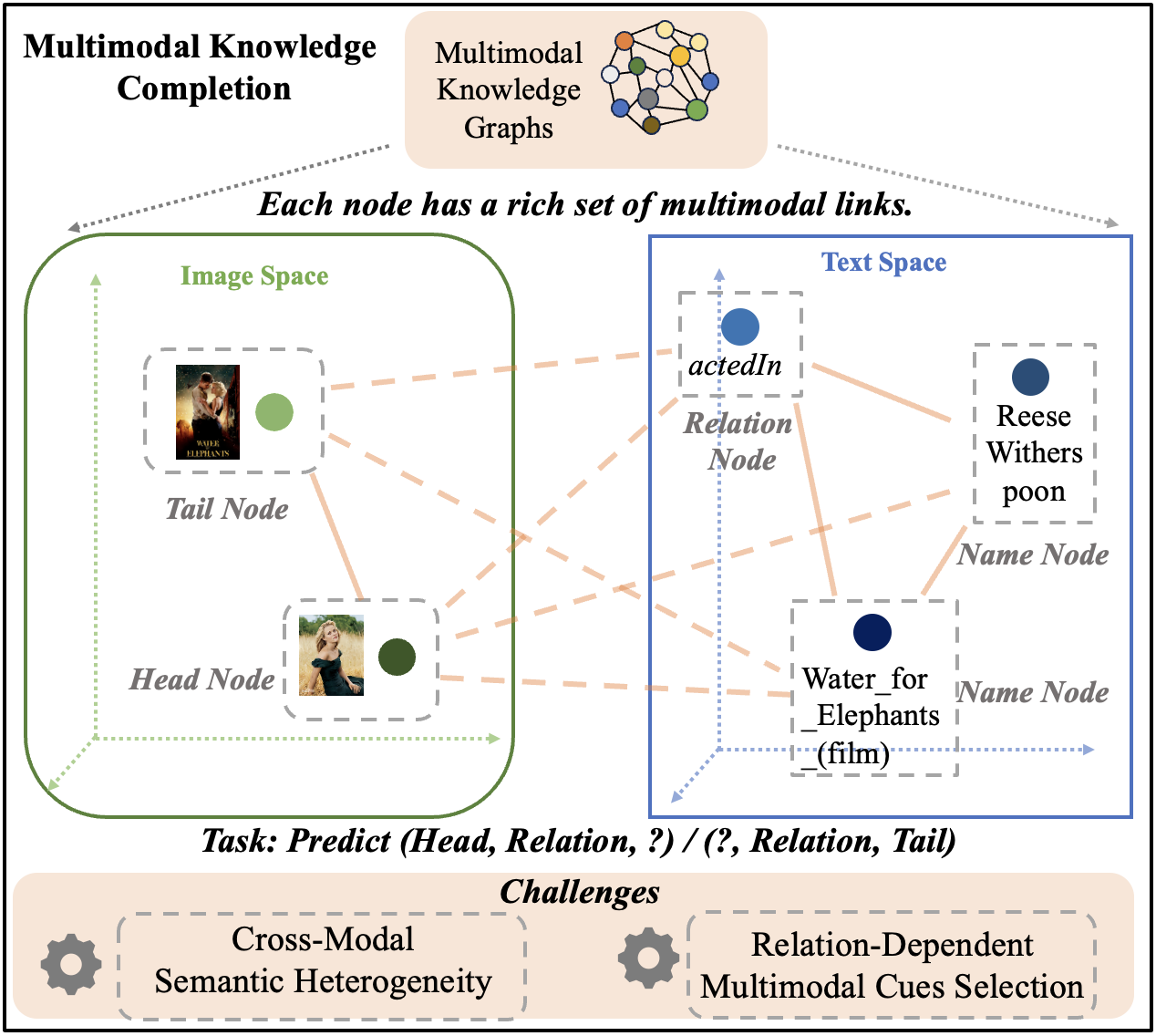}
    \caption{
    Core challenges of MKGC. 
    Structural, textual, and visual features exhibit cross-modal semantic heterogeneity
    and relation-dependent multimodal cue selection. 
    Solid lines indicate intra-modal relations, while dashed lines indicate inter-modal relations. 
    As a result, raw multimodal representations do not form a reliable unified reasoning space for relation-conditioned entity completion.
    }
    \label{fig:challenge}
\end{figure}

Multimodal Knowledge Graph Completion (MKGC) aims to infer missing entities or links in a knowledge graph by jointly leveraging structural relations, textual descriptions, and visual cues. Compared with conventional knowledge graph completion, MKGC is more practical for real-world scenarios, where entities are often associated with rich multimodal attributes beyond symbolic triples. In this setting, the model must answer relation-conditioned queries such as $(h,r,?)$ or $(?,r,t)$ by reasoning over graph structure together with heterogeneous multimodal observations. This makes MKGC substantially more challenging than unimodal completion, since the model must not only identify the correct entity but also ensure that the completed triple remains semantically meaningful and structurally consistent.

Specifically, MKGC is mainly challenged by two coupled difficulties. The first is \textit{cross-modal semantic heterogeneity}: structural, textual, and visual features are produced by different encoders and optimization objectives, causing them to reside in incompatible semantic spaces. Even when describing the same entity, different modalities often emphasize different semantic aspects, resulting in substantial cross-modal semantic gaps. As illustrated in Figure~\ref{fig:challenge}, this makes raw multimodal representations neither directly comparable nor reliably composable as a unified reasoning substrate. The second difficulty is \textit{relation-dependent multimodal cues selection}: different relations often rely on different semantic cues. Some cases are mainly resolved by structural context, while others depend more heavily on textual semantics or visual cues. Thus, effective MKGC requires not only cross-modal alignment, but also relation-aware use of multimodal knowledge before completion.

Existing discriminative methods address these two challenges from different perspectives. Early approaches such as IKRL~\cite{xie2017image} and TransAE~\cite{transAE} primarily alleviate the first challenge by projecting multimodal attributes into a structural embedding space for triple scoring. More recent methods further improve multimodal representation learning for semantically heterogeneous inputs. For example, OTKGE~\cite{OTKGE} addresses cross-modal inconsistency more explicitly by introducing optimal transport to better align multimodal representations, while MyGo~\cite{MyGo} improves multimodal entity modeling through discrete codebook, yielding richer multimodal representations for ranking. In contrast, 
% MoMoK~\cite{zhang2024unleashingpowerimbalancedmodality} focuses more on the second challenge by enhancing multimodal reasoning through relation-aware gating and expert-style modality selection, allowing the model to adaptively emphasize different modalities under different relations.
MoMoK~\cite{zhang2024unleashingpowerimbalancedmodality} mainly addresses the second challenge by expert-style modality selection and adaptively emphasizing different modalities under different relations.
Although these methods have improved multimodal entity ranking, their objectives remain fundamentally discriminative: they estimate compatibility scores over candidate entities, rather than explicitly modeling a generative refinement process for missing entity completion under multimodal uncertainty.

Inspired by the recent success of diffusion models in complex conditional generation~\cite{nichol2021glide, saharia2022imagen,rombach2022high}, researchers have recently begun to explore diffusion-based formulations for multimodal knowledge graph completion. By progressively denoising a latent variable toward a target entity representation, diffusion provides a natural mechanism for iterative refinement under uncertainty, making it particularly well suited to multimodal entity completion. Recent works such as DiffusionCom~\cite{huang2025diffusioncomstructureawaremultimodaldiffusion} and FDM~\cite{FDM} have demonstrated the promise of diffusion for modeling complex relational structures in knowledge graphs. However, existing diffusion-based MKGC methods typically inject raw multimodal features directly into the denoising process, implicitly assuming that the original multimodal feature space is already suitable for diffusion. This design leaves the denoiser to simultaneously handle the two challenges above, namely cross-modal semantic heterogeneity and relation-dependent multimodal cue selection, while also performing structure-aware entity generation. Prior diffusion-based formulations effectively denoise under raw multimodal conditions, thereby entangling multimodal cue selection, semantic alignment, and entity generation in a single process.

\begin{figure}[t]
    \centering
    \includegraphics[width=\linewidth]{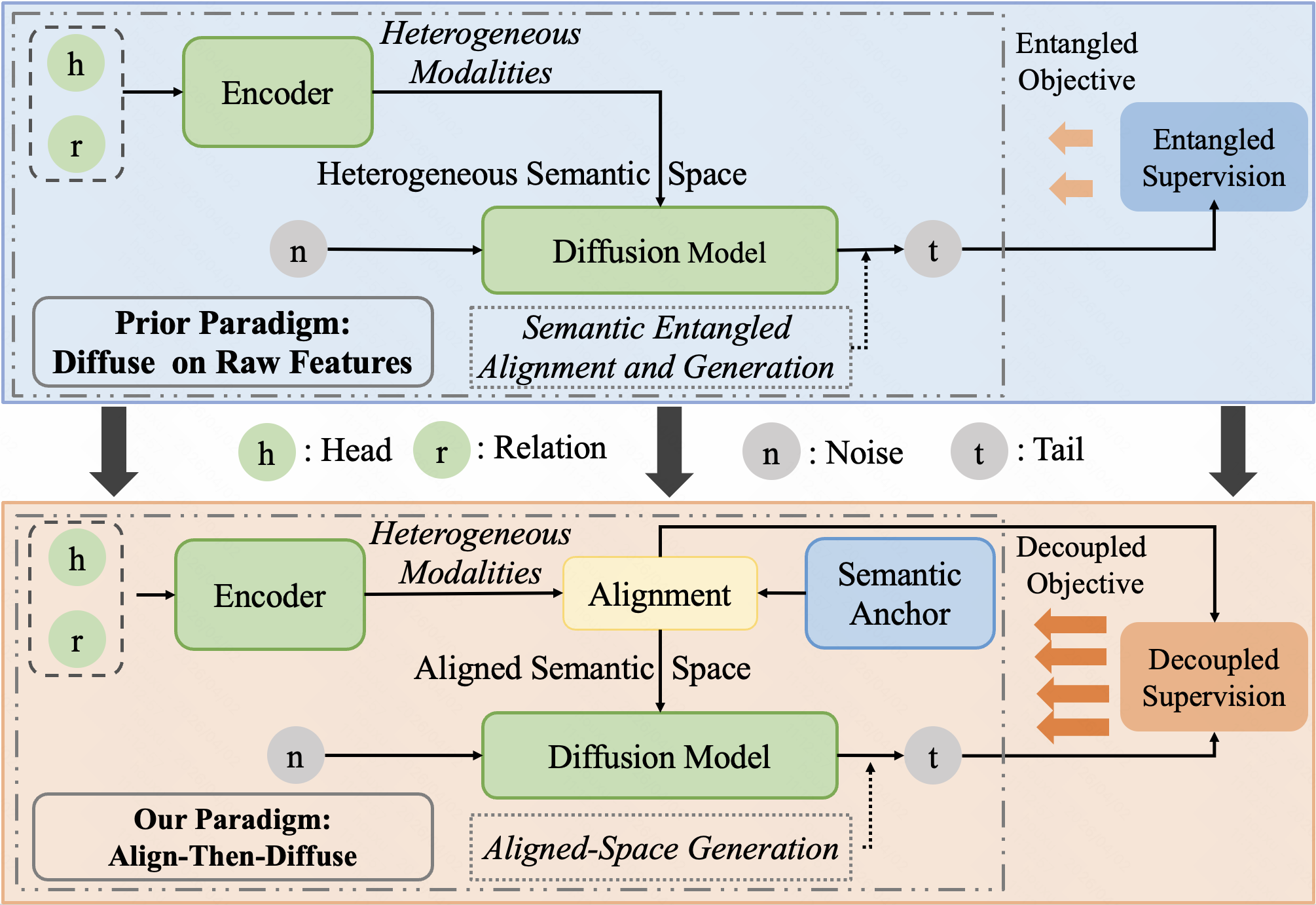}
    \caption{Comparison between the proposed method and existing methods. Existing methods directly denoise under raw heterogeneous multimodal conditions, which entangles semantic alignment with entity generation. In contrast, M$^2$GDT first aligns multimodal representations with an MLLM semantic anchor and then performs structure-aware diffusion generation with decoupled supervision.}
    \label{fig:paradigm}
\end{figure}

Therefore, we argue that effective diffusion for MKGC should be built on a \textit{relation-dependent and semantically aligned} latent space rather than the raw heterogeneous feature space. Figure~\ref{fig:paradigm} contrasts this idea with prior diffusion-based formulations. Based on this intuition, we propose \textbf{M}LLM-\textbf{G}uided \textbf{D}iffusion \textbf{T}ransformer with Relation-Adaptive \textbf{M}ixture-of-Experts(\textbf{M$^2$GDT}), a new framework for MKGC built on an \textit{align-then-diffuse} paradigm. As illustrated in Figure~\ref{fig:paradigm}, M$^2$GDT explicitly addresses the above two challenges in a decoupled manner. First, it performs \textit{Relation-Adaptive Semantic Routing} with a mixture-of-experts (MoE) module, which dynamically selects modality-specific semantic transformation paths under relation conditioning and thus tackles relation-dependent multimodal cue selection. Next, the routed multimodal representations are aligned into a unified semantic space under the guidance of a Multimodal Large Language Model (MLLM), which serves as a semantic anchor with strong cross-modal priors and alleviates semantic heterogeneity across modalities. Finally, diffusion is performed in this aligned latent space with a \textbf{K}nowledge \textbf{G}raph \textbf{D}iffusion \textbf{T}ransformer (\textbf{KGDT}), allowing the denoiser to focus on graph-conditioned entity generation rather than being jointly burdened by multimodal cue selection, alignment, and generation. 
% Experiments validate the effectiveness of M$^2$GDT.
Extensive experiments on three benchmark datasets demonstrate that M$^2$GDT consistently outperforms strong baselines, validating the effectiveness of the proposed paradigm.

Our contributions are summarized as follows:
\begin{itemize}
    \item We propose \textbf{M}LLM-\textbf{G}uided \textbf{D}iffusion \textbf{T}ransformer with Relation-Adaptive \textbf{M}ixture-of-Experts(\textbf{M$^2$GDT}), a novel framework for Multimodal Knowledge Graph Completion built on an \textit{align-then-diffuse} paradigm, which performs diffusion generation in a relation-dependent and semantically aligned latent space rather than directly in raw multimodal feature spaces.
    
    \item We design a \textbf{Relation-Adaptive Semantic Routing Mixture of Experts(RASR-MoE)} module that explicitly addresses relation-dependent multimodal cue selection by dynamically choosing modality-specific semantic transformation paths under relation conditioning, producing cleaner multimodal representations for subsequent MLLM-anchored alignment. We further employ a lightweight auxiliary regularization term to encourage complementary expert behaviors.
    
    \item We develop an MLLM-anchored alignment and diffusion generation framework, in which routed multimodal representations are first calibrated into a unified semantic space to alleviate cross-modal semantic heterogeneity and then refined by the proposed \textbf{K}nowledge \textbf{G}raph \textbf{D}iffusion \textbf{T}ransformer (\textbf{KGDT}) for graph-conditioned denoising. 
    Extensive experiments on three benchmark datasets validate the effectiveness of the proposed framework and demonstrate consistent improvements over strong baselines.
\end{itemize}

\section{Methodology}
\begin{figure*}[t]
    \centering
    \includegraphics[width=0.90\textwidth]{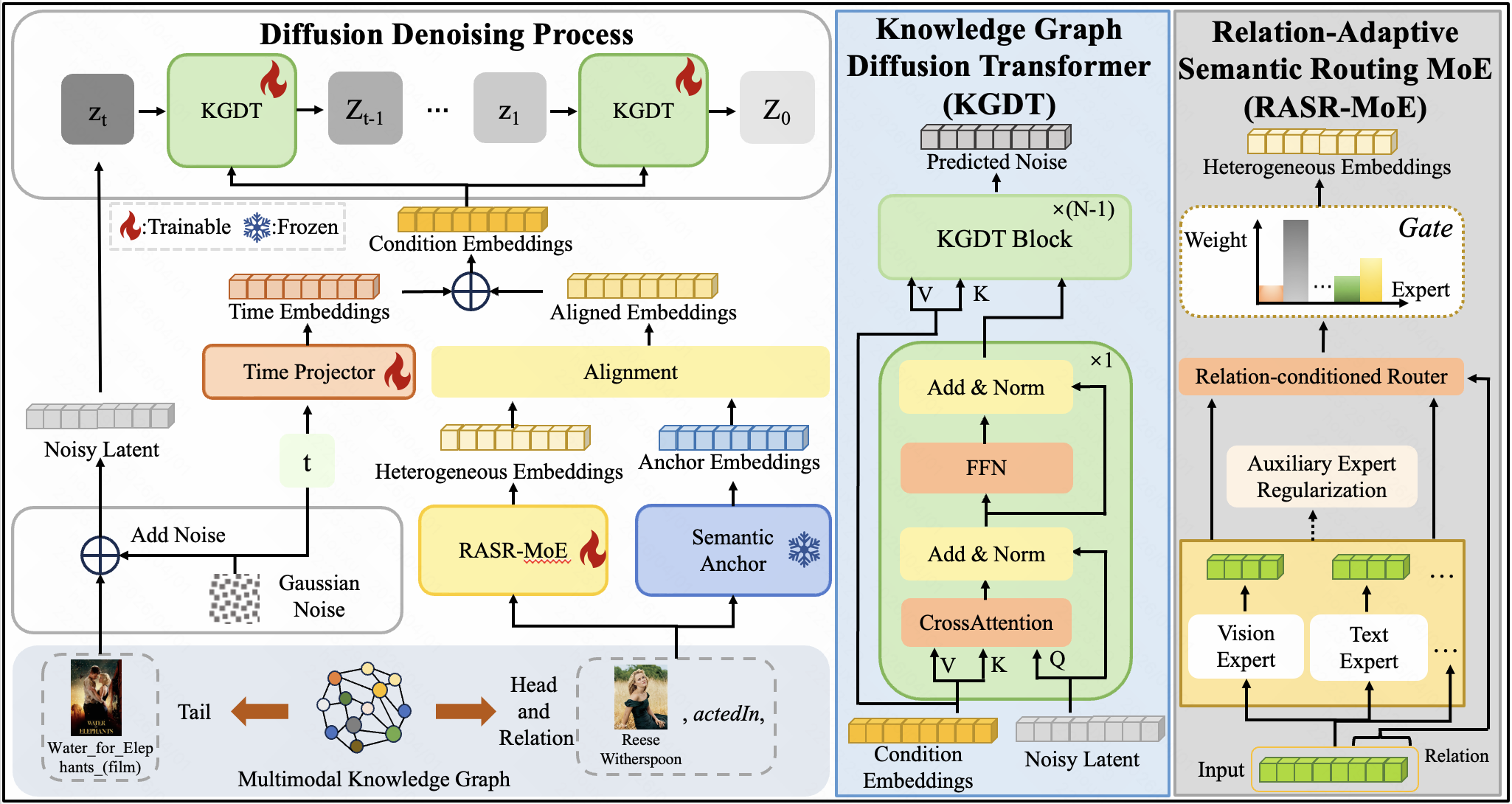}
    \caption{
    Overview of the \textbf{M$^2$GDT} framework. \textbf{RASR-MoE} first performs relation-adaptive semantic routing over heterogeneous multimodal inputs. A frozen semantic anchor then aligns the routed features into calibrated condition embeddings, which are combined with timestep embeddings for diffusion denoising. Starting from a noisy latent, \textbf{KGDT} iteratively predicts noise and refines the latent toward the target entity representation. The right panels show the architectures of \textbf{KGDT} and \textbf{RASR-MoE}.
    }
    \label{fig:framework}
\end{figure*}

\subsection{Problem Formulation}
A Multimodal Knowledge Graph is denoted as
\begin{equation}
\mathcal{G}=(\mathcal{V},\mathcal{R},\mathcal{T},\mathcal{M}),
\end{equation}
where $\mathcal{V}$, $\mathcal{R}$, $\mathcal{T}$, and $\mathcal{M}$ denote the entity set, relation set, factual triples, and multimodal attributes, respectively. Given an incomplete query $(h,r,?)$ or $(?,r,t)$, the goal of MKGC is to infer the missing entity so that the completed triple is both semantically plausible and structurally consistent. Unlike conventional discriminative methods that directly rank candidate entities, diffusion-based methods formulate MKGC as a conditional generation problem in the latent space. Without loss of generality, we focus on tail prediction $(h,r,?)$, while the head prediction case is symmetric. Given the observed entity $h$, relation $r$, and the associated multimodal information, the model generates the missing entity representation $\hat{\mathbf{z}}_0$ through a denoising diffusion process, which is then matched against candidate entity embeddings for final ranking.

\subsection{Relation-Adaptive Semantic Routing Mixture-of-Experts}
A key assumption of M$^2$GDT is that diffusion should be conditioned on semantically calibrated multimodal representations rather than uniformly fused raw features. In multimodal knowledge graphs, different relations often depend on different semantic information. Some relations are mainly determined by structural context, while others rely more on textual or visual cues. Therefore, using a fixed multimodal encoding path for all queries may introduce unnecessary semantic interference before alignment.

To address this issue, we introduce a \textbf{Relation-Adaptive Semantic Routing Mixture-of-Experts} (\textbf{RASR-MoE}) module, which performs relation-conditioned expert routing before anchor-guided alignment. Its goal is not generic multimodal fusion, but selecting relation-dependent semantic transformation paths that produce cleaner representations for subsequent MLLM-guided alignment.

Given a query $(h,r,?)$, we first retrieve the raw structural, visual, and textual features of the observed entity $h$:
\begin{align}
    \mathbf{x}_s &= \mathrm{EntEmb}(h)\in\mathbb{R}^{B\times d_s},\\
    \mathbf{x}_v &= \mathrm{ImgEmb}(h)\in\mathbb{R}^{B\times d_v},\\
    \mathbf{x}_t &= \mathrm{TxtEmb}(h)\in\mathbb{R}^{B\times d_t},
\end{align}
where $B$ is the batch size, and $\mathbf{r}\in\mathbb{R}^{B\times d_r}$ denotes the relation embedding.

For each modality $k\in\{s,v,t\}$, we define a bank of $M$ experts
\[
\{\mathrm{E}_{k}^{(1)},\mathrm{E}_{k}^{(2)},\dots,\mathrm{E}_{k}^{(M)}\}.
\]
A relation-conditioned router computes the expert weights:
\begin{align}
    \tilde{\mathbf{g}}_k
    =
    \mathrm{softmax}
    \big(
    \mathrm{MLP}_k([\mathbf{x}_k;\mathbf{r}]) + \boldsymbol{\delta}_k
    \big),
    \qquad
    \boldsymbol{\delta}_k \sim \mathcal{N}(\mathbf{0},\Sigma_k),
\end{align}
where $\Sigma_k$ is predicted from the current modality representation. This perturbation improves routing robustness and avoids over-confident expert selection.

Each expert produces a relation-aware transformation
\begin{align}
    \mathbf{u}_{k}^{(m)}
    =
    \mathrm{E}_{k}^{(m)}(\mathbf{x}_k,\mathbf{r}),
    \qquad m=1,\dots,M,
\end{align}
and the routed modality representation is obtained by
\begin{align}
    \tilde{\mathbf{e}}_k
    =
    \sum_{m=1}^{M}
    \tilde{\mathbf{g}}_{k}^{(m)} \odot \mathbf{u}_{k}^{(m)},
    \qquad k\in\{s,v,t\}.
\end{align}

The routed features are then fused into a multimodal entity representation. Concretely, we first project the routed modality features into a shared latent space:
\begin{align}
    \mathbf{e}_k = \mathbf{W}_k \tilde{\mathbf{e}}_k + \mathbf{b}_k,
    \qquad k\in\{s,v,t\},
\end{align}
where $\mathbf{W}_k\in\mathbb{R}^{d_f\times d_k}$ and $\mathbf{b}_k\in\mathbb{R}^{d_f}$ are learnable modality-specific projection parameters. The fused entity representation is then obtained by a weighted sum:
\begin{align}
    \mathbf{e}_f
    =
    \mathbf{e}_s
    +
    \mathbf{e}_v
    +
    \mathbf{e}_t,
\end{align}

In this way, RASR-MoE performs relation-aware semantic routing before alignment, while the subsequent lightweight fusion aggregates the routed modality features into unified multimodal representations for the \textit{align-then-diffuse} process. To preserve task-relevant discriminative capability in the RASR-MoE encoder, we further apply a supervision term to the structural, visual, textual, and fused branches. Following the encoder design, these branches produce prediction scores over all candidate entities:
\begin{align}
    \mathbf{p}_s,\mathbf{p}_v,\mathbf{p}_t,\mathbf{p}_f \in \mathbb{R}^{B\times N},
\end{align}
where $N$ is the number of entities. The corresponding auxiliary supervision loss is defined as
\begin{align}
    \mathcal{L}_{\mathrm{moe}}
    =
    \mathrm{BCE}(\mathbf{p}_s,\mathbf{y})
    +
    \mathrm{BCE}(\mathbf{p}_v,\mathbf{y})
    +
    \mathrm{BCE}(\mathbf{p}_t,\mathbf{y})
    +
    \mathrm{BCE}(\mathbf{p}_f,\mathbf{y}),
\end{align}
where $\mathbf{y}\in\mathbb{R}^{B\times N}$ is the target label matrix. Here, $\mathrm{BCE}(\cdot,\cdot)$ denotes the binary cross-entropy loss, which supervises each branch to assign higher scores to valid target entities and thus preserves task-relevant discriminative signals in the RASR-MoE encoder.

% \subsubsection{Auxiliary Expert Regularization}
To reduce redundancy among experts, we further introduce a lightweight auxiliary regularization term inspired by mutual-information-based expert disentanglement. This term encourages different experts within the same modality to capture complementary relation-dependent semantic transformations rather than collapsing to similar mappings:
\begin{align}
    \mathcal{L}_{\mathrm{reg}}
    =
    \sum_{k\in\{s,v,t\}}
    \mathcal{L}_{\mathrm{club}}^{(k)},
\end{align}
where $\mathcal{L}_{\mathrm{club}}^{(k)}$ is a lightweight CLUB-style regularizer for modality $k$~\cite{zhang2024unleashingpowerimbalancedmodality}. Since this term only serves as an auxiliary objective, we keep its weight small in the overall training loss.

\subsection{MLLM-Anchored Representation Alignment}

\subsubsection{Frozen Anchor Construction}
For the same batch of entities, we concatenate the raw structural, visual, and textual inputs:
\begin{align}
    \mathbf{x}_{\mathrm{cat}}
    =
    [\mathbf{x}_s;\mathbf{x}_v;\mathbf{x}_t]
    \in
    \mathbb{R}^{B\times (d_s+d_v+d_t)}.
\end{align}
A learnable projection maps them into the hidden dimension of the frozen Qwen3-VL language backbone:
\begin{align}
    \mathbf{x}_{\mathrm{anchor}}
    =
    \mathrm{Proj}_{\mathrm{in}}(\mathbf{x}_{\mathrm{cat}})
    \in
    \mathbb{R}^{B\times d_q},
\end{align}
where $d_q=4096$ in our implementation.

The projected features are treated as continuous input embeddings and fed into the frozen backbone:
\begin{align}
    \mathbf{h}^{\mathrm{anchor}}
    =
    \mathrm{Qwen}_{\mathrm{frozen}}(\mathbf{x}_{\mathrm{anchor}})
    \in
    \mathbb{R}^{B\times d_q}.
\end{align}
All parameters of the Qwen backbone are frozen throughout training.

\subsubsection{Modality-specific Anchor Outputs}
The frozen hidden states are projected into modality-specific spaces to obtain anchor features with the same output dimensions as the MoE encoder:
\begin{align}
    \mathbf{a}_s &= \mathrm{Proj}^{a}_{s}(\mathbf{h}^{\mathrm{anchor}})
    \in\mathbb{R}^{B\times d_s},\\
    \mathbf{a}_v &= \mathrm{Proj}^{a}_{v}(\mathbf{h}^{\mathrm{anchor}})
    \in\mathbb{R}^{B\times d_v},\\
    \mathbf{a}_t &= \mathrm{Proj}^{a}_{t}(\mathbf{h}^{\mathrm{anchor}})
    \in\mathbb{R}^{B\times d_t},\\
    \mathbf{a}_f &= \mathrm{Proj}^{a}_{f}(\mathbf{h}^{\mathrm{anchor}})
    \in\mathbb{R}^{B\times d_f}.
\end{align}
Therefore, the MoE encoder outputs and the anchor outputs have matched tensor shapes and can be aligned directly.

\subsubsection{Feature-level Alignment}
We align the MoE output features $\mathbf{e}_s,\mathbf{e}_v,\mathbf{e}_t,\mathbf{e}_f$ with the corresponding anchor features $\mathbf{a}_s,\mathbf{a}_v,\mathbf{a}_t,\mathbf{a}_f$.

For the structural, visual, and textual branches, we use cosine-similarity-based alignment:
\begin{align}
    \mathcal{L}_{k}
    =
    1
    -
    \frac{1}{B}
    \sum_{n=1}^{B}
    \cos\!\left(\mathbf{e}_{k,n},\mathbf{a}_{k,n}\right),
    \qquad k\in\{s,v,t\},
\end{align}
where $\mathbf{e}_{k,n}$ and $\mathbf{a}_{k,n}$ denote the $n$-th sample in the batch.

For the fused branch, we use a temperature-scaled KL matching loss:
\begin{align}
    \mathbf{p}_{f,n}
    &=
    \mathrm{softmax}(\mathbf{e}_{f,n}/T),\\
    \mathbf{q}_{f,n}
    &=
    \mathrm{softmax}(\mathbf{a}_{f,n}/T),
\end{align}
\begin{align}
    \mathcal{L}_{f}
    =
    \frac{1}{B}
    \sum_{n=1}^{B}
    \mathrm{KL}\!\left(
    \mathbf{p}_{f,n}\,\|\,\mathbf{q}_{f,n}
    \right),
\end{align}
where $T$ is the temperature.

The total alignment loss is defined as
\begin{align}
    \mathcal{L}_{\mathrm{align}}
    =
    \lambda_k \mathcal{L}_k
    +
    \lambda_f \mathcal{L}_f.
\end{align}
In the current implementation, the strongest alignment weights are assigned to the structural and visual branches, which are heuristically set to 1.0, while the textual and fused branches are used as auxiliary regularizers.

\subsection{Knowledge Graph Diffusion Transformer}
With the multimodal condition now aligned in a unified semantic space, M$^2$GDT performs the \emph{diffuse} stage by generating the missing entity embedding through a conditional denoising diffusion process. We instantiate the denoiser as a \textbf{K}nowledge \textbf{G}raph \textbf{D}iffusion \textbf{T}ransformer (KGDT), whose architecture is shown in Figure~\ref{fig:framework}. Compared with linear denoisers, a transformer\cite{vaswani2017attention} backbone is better suited for modeling long-range relational dependencies and condition-dependent interactions in knowledge graphs.

\begin{figure}[t]
    \centering
    \includegraphics[width=1.0\linewidth]{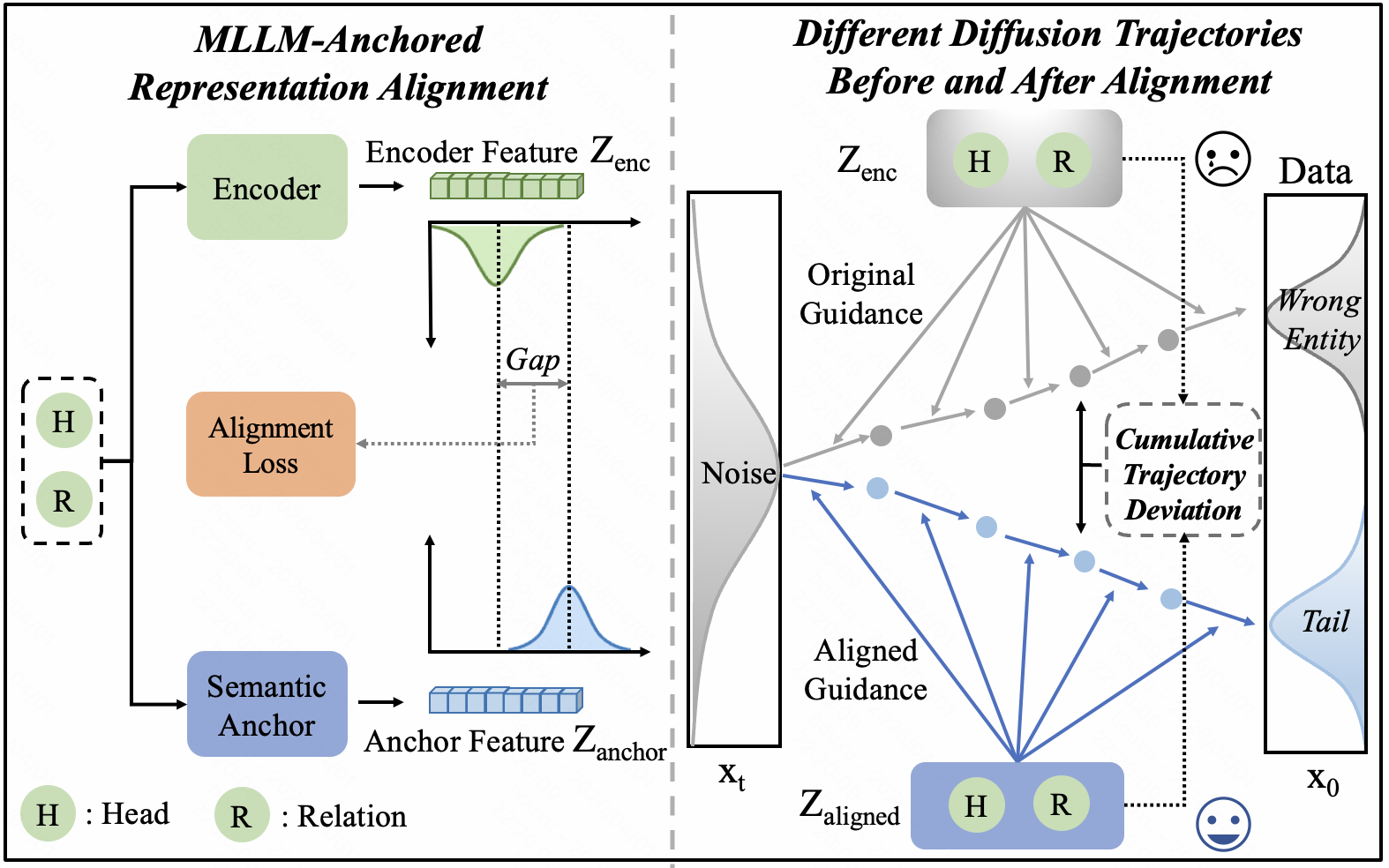}
    \caption{
    Illustration of the proposed anchor-guided alignment mechanism.
    The encoder outputs are aligned to frozen anchor references through feature-level similarity losses, producing a semantically calibrated condition for diffusion.
    }
    \label{fig:alignment}
\end{figure}

\subsubsection{Diffusion Condition under Joint Alignment}

The anchor branch does not explicitly replace the encoder outputs with pre-aligned features. Instead, the alignment loss $\mathcal{L}_{\mathrm{align}}$ is optimized jointly with the diffusion objective during training, so that the encoder representations are progressively calibrated toward the frozen anchor references. Therefore, the diffusion model is always conditioned on the current encoder features, while semantic alignment is enforced as a simultaneous training constraint.

Concretely, at each training step, the routed modality representations produced by the RASR-MoE encoder are
\begin{align}
    \mathbf{e}_s\in\mathbb{R}^{B\times d_f},\quad
    \mathbf{e}_v\in\mathbb{R}^{B\times d_f},\quad
    \mathbf{e}_t\in\mathbb{R}^{B\times d_f}.
\end{align}
These features are then combined by element-wise addition to form the entity-side semantic condition:
\begin{align}
    \mathbf{e}_f
    =
    \mathbf{e}_s
    +
    \mathbf{e}_v
    +
    \mathbf{e}_t \in\mathbb{R}^{B\times d_f},
\end{align}

For each diffusion step $t$, a learnable time projector maps the timestep index to a time embedding:
\begin{align}
    \bm{\tau}_{t}
    =
    \mathrm{Proj}_{\mathrm{time}}(t),
\end{align}
which is broadcast to all samples in the batch. The final condition embedding used by the denoiser is then
\begin{align}
    \mathbf{c}_{t}
    =
    \mathbf{e}_{f}
    +
    \bm{\tau}_{t}.
\end{align}

In this way, semantic alignment and diffusion conditioning are learned jointly rather than sequentially: the denoiser always receives the current encoder condition, while the anchor-guided loss continuously regularizes this condition toward a semantically coherent latent space throughout training.

\subsubsection{Forward Diffusion}
Let $\mathbf{z}_0$ denote the target embedding of the missing entity. Following DDPM~\cite{nichol2021improveddenoisingdiffusionprobabilistic}, the forward diffusion process gradually perturbs $\mathbf{z}_0$ with Gaussian noise:
\begin{align}
    \mathbf{z}_t
    =
    \sqrt{\bar{\alpha}_t}\mathbf{z}_0
    +
    \sqrt{1-\bar{\alpha}_t}\,\bm{\epsilon},
    \qquad
    \bm{\epsilon}\sim\mathcal{N}(\mathbf{0},\mathbf{I}),
\end{align}
where $\bar{\alpha}_t=\prod_{i=1}^{t}\alpha_i$ and $\alpha_t=1-\beta_t$ is determined by the variance schedule $\{\beta_t\}_{t=1}^{T}$.

\begin{table*}[tbp]
\centering
\caption{MKGC performance comparison on MKG-W, MKG-Y and DB15K datasets. The best results are marked in bold, the second-best results are underlined.}
\label{tab:main_results}
\begin{tabular}{lcccccccccccc}
\toprule
\textbf{Methods} & \multicolumn{4}{c}{\textbf{MKG-W}} & \multicolumn{4}{c}{\textbf{MKG-Y}} & \multicolumn{4}{c}{\textbf{DB15K}} \\
\cmidrule(r){2-5} \cmidrule(r){6-9} \cmidrule(l){10-13}
 & \textbf{MRR} & \textbf{H@1} & \textbf{H@3} & \textbf{H@10} & \textbf{MRR} & \textbf{H@1} & \textbf{H@3} & \textbf{H@10} & \textbf{MRR} & \textbf{H@1} & \textbf{H@3} & \textbf{H@10} \\
\midrule

\multicolumn{13}{c}{\textit{Unimodal KG Completion}} \\
\addlinespace[2pt]

TransE & 29.19 & 21.06 & 33.20 & 44.23 & 30.73 & 23.45 & 35.18 & 43.37 & 24.86 & 12.78 & 31.48 & 47.07 \\
DistMult & 20.99 & 15.93 & 22.28 & 30.86 & 28.71 & 22.26 & 27.80 & 35.95 & 23.03 & 14.78 & 26.28 & 39.59 \\
ComplEx & 24.93 & 19.09 & 26.69 & 36.73 & 25.04 & 19.33 & 32.12 & 40.93 & 27.48 & 18.37 & 31.57 & 45.37 \\
RotatE & 33.67 & 26.80 & 36.68 & 46.73 & 34.95 & 29.10 & 38.35 & 45.30 & 29.28 & 17.87 & 36.12 & 49.66 \\
GC-OTE & 33.92 & 26.55 & 35.96 & 46.05 & 32.95 & 26.77 & 36.44 & 44.08 & 31.85 & 22.11 & 36.52 & 51.18 \\
TuckER & 30.40 & 24.40 & 32.90 & 41.30 & 37.10 & 34.60 & 38.40 & 42.90 & 30.10 & 20.50 & 35.70 & 46.80 \\

\addlinespace[4pt]
\multicolumn{13}{c}{\textit{Multimodal KG Completion}} \\
\addlinespace[2pt]

IKRL & 32.36 & 26.11 & 34.75 & 44.07 & 33.22 & 30.37 & 34.28 & 38.26 & 26.82 & 14.09 & 34.93 & 49.09 \\
TBKGC & 31.48 & 25.31 & 33.98 & 43.24 & 33.99 & 30.47 & 35.27 & 40.07 & 28.40 & 15.61 & 37.03 & 49.86 \\
TransAE & 30.00 & 21.23 & 34.91 & 44.72 & 28.10 & 25.31 & 29.10 & 33.03 & 28.09 & 21.25 & 31.17 & 41.17 \\
MMKRL & 30.10 & 22.16 & 34.90 & 44.69 & 36.81 & 31.66 & 39.79 & \underline{45.31} & 26.81 & 13.85 & 35.07 & 49.39 \\
RSME & 29.23 & 23.36 & 31.97 & 40.43 & 34.44 & 31.78 & 36.07 & 39.09 & 29.76 & 24.15 & 32.12 & 40.29 \\
VBKGC & 30.61 & 24.91 & 33.01 & 40.88 & 37.04 & 33.76 & 38.75 & 42.30 & 30.61 & 19.75 & 37.18 & 49.44 \\
OTKGE & 34.36 & 28.85 & 36.25 & 44.88 & 35.51 & 31.97 & 37.18 & 41.38 & 23.86 & 14.85 & 25.89 & 34.83 \\
IMF & 32.58 & 27.77 & 36.06 & 45.44 & 35.79 & 32.95 & 37.10 & 40.60 & 32.25 & 24.20 & 36.06 & 48.19 \\
AdaMF-MAT & 34.27 & 27.21 & 37.86 & 47.21 & 38.06 & 33.49 & 40.44 & \textbf{45.48} & 32.51 & 21.31 & 39.67 & 51.68 \\
VISTA & 32.91 & 26.12 & 35.38 & 45.61 & 30.45 & 24.87 & 32.40 & 41.50 & 30.42 & 22.49 & 33.56 & 45.94 \\
MyGo & 36.10 & 29.78 & 38.54 & \underline{47.75} & \underline{38.44} & 35.01 & 39.84 & 44.19 & 37.72 & 30.08 & 41.26 & 52.21 \\
MoMoK & 35.89 & 30.38 & 37.54 & 46.13 & 37.91 & 35.09 & 39.20 & 43.20 & 39.57 & \underline{32.38} & 43.45 & 54.14 \\
MCKGC & \underline{36.88} & \underline{31.32} & \underline{38.92} & 47.43 & 38.06 & \underline{35.49} & \underline{40.57} & 45.21 & \underline{39.79} & 31.92 & \underline{43.80} & \underline{54.66} \\

\addlinespace[2pt]
\midrule

\textbf{M$^2$GDT (Ours)} & \textbf{37.77} & \textbf{32.18} & \textbf{39.83} & \textbf{48.15} & \textbf{39.25} & \textbf{36.12} & \textbf{40.96} & 45.10 & \textbf{40.54} & \textbf{32.60} & \textbf{44.49} & \textbf{55.48} \\

\midrule

\textit{\textbf{Gain vs. SOTA}} & \textit{2.41\% $\uparrow$} & \textit{2.75\% $\uparrow$} & \textit{2.34\% $\uparrow$} & \textit{0.84\% $\uparrow$} & \textit{3.13\% $\uparrow$} & \textit{1.78\% $\uparrow$} & \textit{0.96\% $\uparrow$} & \textit{-0.84\% $\downarrow$} & \textit{1.88\% $\uparrow$} & \textit{0.68\% $\uparrow$} & \textit{1.58\% $\uparrow$} & \textit{1.50\% $\uparrow$} \\

\bottomrule
\end{tabular}
\end{table*}

\subsubsection{Conditional Reverse Denoising}
The reverse denoising process is parameterized by KGDT:
\begin{align}
    \hat{\bm{\epsilon}} = \mathrm{KGDT}(\mathbf{z}_t,\mathbf{c}_t;\Theta),
\end{align}
where $\Theta$ denotes the learnable parameters of the transformer. As shown in Figure~\ref{fig:framework}, the first KGDT block injects $\mathbf{c}_t$ into the noisy latent through cross-attention, and the subsequent transformer blocks iteratively refine the latent representation through stacked attention and feed-forward transformations. In this way, each denoising step is guided by semantically aligned multimodal information rather than by raw heterogeneous features.

Given the predicted noise, the reverse transition follows the standard DDPM update:
\begin{align}
    \mathbf{z}_{t-1}
    =
    \frac{1}{\sqrt{\alpha_t}}
    \left(
    \mathbf{z}_t
    -
    \frac{1-\alpha_t}{\sqrt{1-\bar{\alpha}_t}}
    \hat{\bm{\epsilon}}
    \right)
    +
    \sigma_t \bm{\eta},
    \qquad
    \bm{\eta}\sim\mathcal{N}(\mathbf{0},\mathbf{I}),
\end{align}
where $\sigma_t$ is determined by the diffusion schedule.

\subsection{Training and Inference}

\subsubsection{Training Objective}

The full M$^2$GDT framework is trained with four objectives: an auxiliary supervision term for the RASR-MoE, a lightweight auxiliary expert regularization term for the RASR-MoE, an anchor-guided semantic alignment objective, and a conditional diffusion denoising objective.

First, the auxiliary supervision loss $\mathcal{L}_{\mathrm{moe}}$ preserves task-relevant discriminative signals in the RASR-MoE encoder, while the auxiliary regularization term $\mathcal{L}_{\mathrm{reg}}$ encourages different experts to capture complementary relation-dependent semantic transformations.

Second, the alignment loss $\mathcal{L}_{\mathrm{align}}$ encourages the routed multimodal features to match the frozen anchor references in a semantically coherent latent space.

Third, KGDT is trained with a standard denoising objective in the aligned latent space. Given a clean target entity embedding $\mathbf{z}_0$, a sampled timestep $t$, and Gaussian noise $\bm{\epsilon}\sim\mathcal{N}(\mathbf{0},\mathbf{I})$, the diffusion loss is formulated as
\begin{align}
    \mathcal{L}_{\mathrm{diff}}
    =
    \mathbb{E}_{\mathbf{z}_0,t,\bm{\epsilon}}
    \left[
    \left\|
    \mathrm{KGDT}(\mathbf{z}_t,\mathbf{c}_t)-\bm{\epsilon}
    \right\|_2^2
    \right].
\end{align}

The overall training objective is
\begin{align}
    \mathcal{L}_{\mathrm{total}}
    =
    \lambda_{1}\mathcal{L}_{\mathrm{moe}}
    +
    \lambda_{2}\mathcal{L}_{\mathrm{reg}}
    +
    \lambda_{3}\mathcal{L}_{\mathrm{align}}
    +
    \lambda_{4}\mathcal{L}_{\mathrm{diff}}.
\end{align}
Here, $\lambda_{1}$, $\lambda_{2}$, $\lambda_{3}$, and $\lambda_{4}$ control the relative importance of auxiliary supervision, expert regularization, semantic alignment, and diffusion denoising, respectively. In the final model, we set $\lambda_{1}=1.0$, $\lambda_{2}=0.0001$, $\lambda_{3}=0.01$, and $\lambda_{4}=0.1$. This setting keeps the expert regularization lightweight, while balancing semantic alignment and diffusion denoising during joint optimization.

\subsubsection{Inference}
During inference, M$^2$GDT starts from Gaussian noise $\mathbf{z}_T\sim\mathcal{N}(\mathbf{0},\mathbf{I})$ and iteratively applies the reverse denoising process conditioned on the aligned multimodal representation. After $T$ steps, the model obtains the predicted entity embedding $\hat{\mathbf{z}}_0$, which is then compared with candidate entity embeddings for ranking-based completion. In this way, semantic alignment is handled before generation, while KGDT focuses on structure-aware denoising in the aligned latent space.

\section{Experiments}
\subsection{Experimental Setup}
\noindent\textbf{Datasets.}
We evaluate M$^2$GDT on three widely used multimodal knowledge graph completion benchmarks: MKG-W, MKG-Y~\cite{MKG-YW}, and DB15K~\cite{liu2019mmkg}. These datasets contain relational triples together with textual descriptions and visual attributes for entities, making them suitable for evaluating multimodal reasoning and completion.

\noindent\textbf{Task and Evaluation Metrics.}
We follow the standard MKGC protocol, in which the task is to predict the missing head entity for $(?,r,t)$ or the missing tail entity for $(h,r,?)$. For each query, all entities in the dataset are considered as candidates and ranked according to the predicted score. We report Mean Reciprocal Rank (MRR), Hits@1, Hits@3, and Hits@10, for which higher values indicate better performance.

\noindent\textbf{Baselines.}
We compare M$^2$GDT with 19 representative baselines, including 6 unimodal KGC methods and 12 multimodal KGC methods. 
The unimodal baselines include TransE~\cite{bordes2013translating}, DistMult~\cite{DistMult}, ComplEx~\cite{ComplEx}, RotatE~\cite{sun2019rotate}, GC-OTE~\cite{tang2020orthogonal}, and TuckER~\cite{balazevic2019tucker}. 
The multimodal baselines include IKRL~\cite{xie2017image}, TBKGC~\cite{TBKGC}, TransAE~\cite{transAE}, MMKRL~\cite{MMKRL}, RSME~\cite{RSME}, VBKGC~\cite{VBKGC}, OTKGE~\cite{OTKGE}, IMF~\cite{li2023imf}, AdaMF-MAT~\cite{zhang2024unleashingpowerimbalancedmodality}, VISTA~\cite{lee-etal-2023-vista}, MyGo~\cite{MyGo}, MoMoK~\cite{zhang2024unleashingpowerimbalancedmodality}, and MCKGC~\cite{mckgc}.

\noindent\textbf{Implementation Details.}
Our model is implemented in PyTorch 2.6.0 and trained on a single NVIDIA V100 GPU with 32GB memory. We use AdamW with an initial learning rate of $1\times10^{-4}$ and a weight decay of $5\times10^{-5}$. The batch size is set to 128 and the maximum number of training epochs is 3000. Textual features are initialized with BERT-base~\cite{bert}, visual features are extracted by ResNet-50~\cite{ressnet} pretrained on ImageNet~\cite{imagenet}, and structural embeddings are initialized from TuckER. The number of KGDT blocks is set to 3, and the number of diffusion steps is set to 100.

\subsection{Comparison with State-of-the-art}
\begin{figure}[t]
    \centering
    \includegraphics[width=0.98\linewidth]{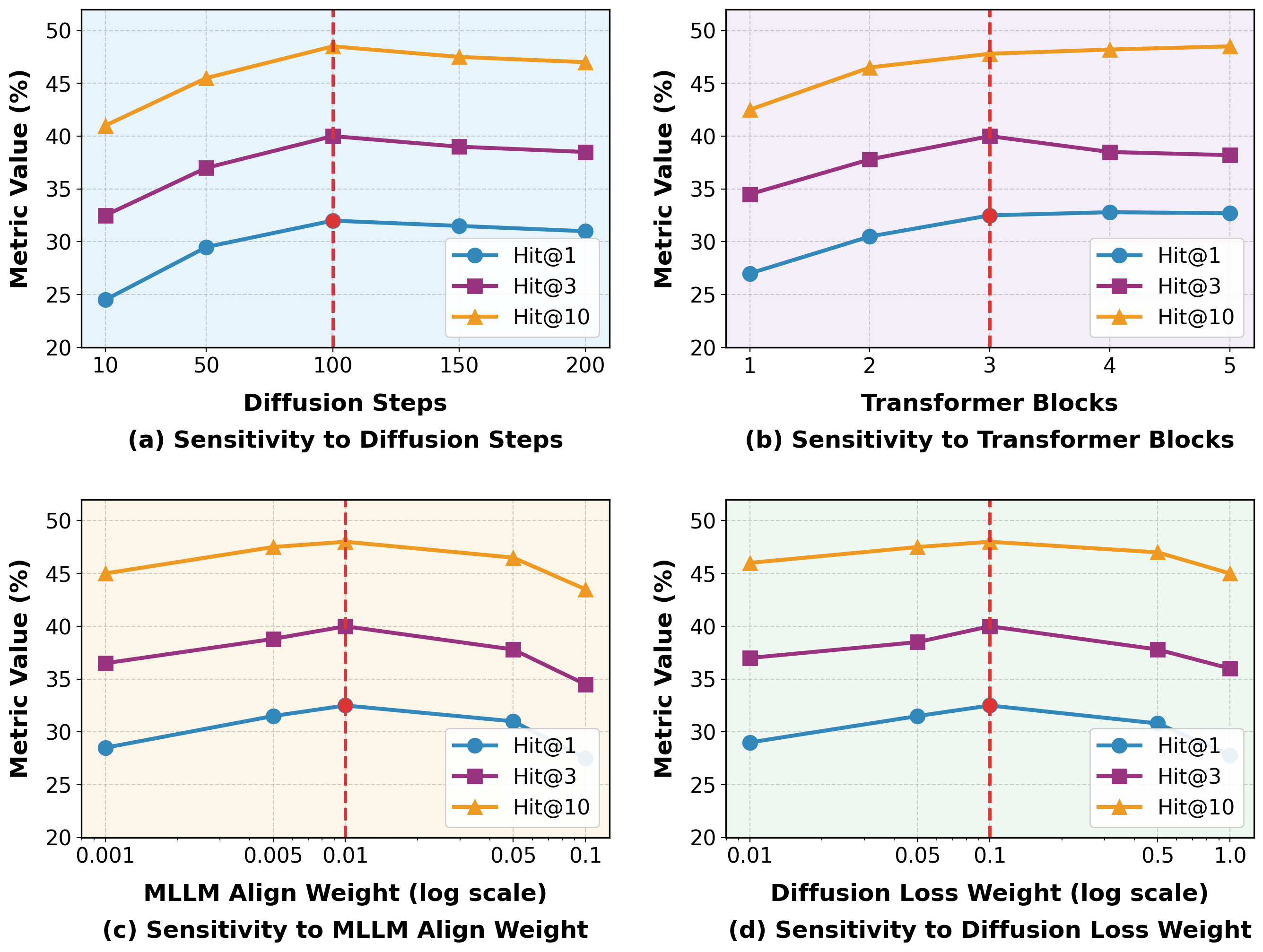}
    \caption{Sensitivity analysis of M$^2$GDT with respect to four hyperparameters on the MKG-W dataset: (a) diffusion steps, (b) number of Transformer blocks, (c) MLLM alignment weight, and (d) diffusion loss weight.} 
    \label{fig:sensitivity}
\end{figure}
Table~\ref{tab:main_results} reports the main results on MKG-W, MKG-Y, and DB15K. Overall, M$^2$GDT achieves the best MRR on all three datasets and also obtains the strongest Hits@1 and Hits@3 in nearly all cases, indicating clear improvements in top-ranked prediction quality. Although M$^2$GDT is not the best method on MKG-Y Hits@10, it remains competitive on this metric while outperforming prior methods on the more discriminative ranking measures. These results show that M$^2$GDT improves not only overall ranking performance, but more importantly the precision of the highest-confidence predictions, which is particularly critical for entity completion.

Compared with unimodal KGC baselines, M$^2$GDT delivers substantial gains by incorporating multimodal cues into a relation-dependent and semantically aligned generative framework. For example, relative to RotatE, one of the strongest unimodal baselines, M$^2$GDT improves MRR from 33.67 to 37.77 on MKG-W, from 34.95 to 39.25 on MKG-Y, and from 29.28 to 40.54 on DB15K. This consistent margin suggests that purely structural reasoning is insufficient for multimodal entity completion, especially when textual and visual cues provide complementary semantics that cannot be captured by structure-only models.

Compared with strong multimodal baselines, M$^2$GDT also shows clear and consistent advantages. On MKG-W, M$^2$GDT improves MRR from 36.88 to 37.77 over the strongest previous result, while also achieving the best Hits@1 and Hits@3. On MKG-Y, M$^2$GDT raises MRR from 38.44 to 39.25 over the best prior baseline, and further achieves the strongest Hits@1 and Hits@3. On DB15K, M$^2$GDT improves MRR from 39.79 to 40.54 and achieves the best performance on all reported metrics. These gains are particularly meaningful because the compared baselines, including MyGo, MoMoK, and MCKGC, are already strong multimodal methods with competitive fusion or representation mechanisms.

More importantly, the empirical advantage of M$^2$GDT is consistent with our method design. Relative to discriminative methods such as MoMoK, which mainly improve multimodal reasoning through relation-aware expert selection, M$^2$GDT further introduces relation-adaptive semantic routing before alignment and performs diffusion-based generation in an MLLM-calibrated latent space. This design allows the model to first select relation-dependent multimodal cues, then reduce cross-modal semantic inconsistency, and finally focus on structure-aware entity generation. Therefore, the gain of M$^2$GDT does not only come from using more multimodal signals, but also from the coordinated effect of \textit{relation-adaptive routing}, \textit{semantic alignment}, and \textit{aligned-space diffusion generation}.

\begin{table}[tbp]
    \centering
    \caption{Ablation study results on the MKG-W dataset.}
    \label{tab:ablation}
    \begin{tabular}{@{}p{3.1cm}cccc@{}}
        \toprule
        \textbf{Variant} & \textbf{MRR} & \textbf{H@1} & \textbf{H@3} & \textbf{H@10} \\
        \midrule
        \textbf{Full M$^2$GDT} & \textbf{37.77} & \textbf{32.18} & \textbf{39.83} & \textbf{48.15} \\
        \hline
        w/o Text Branch & 36.21 & 30.77 & 38.15 & 47.02 \\
        w/o Image Branch & 36.89 & 31.40 & 38.94 & 47.56 \\
        w/o Structure Branch & 35.12 & 29.44 & 37.28 & 46.33 \\
        w/o Semantic Anchor & 33.45 & 27.89 & 35.67 & 43.98 \\
        w/o RASR-MoE & 30.88 & 23.12 & 29.74 & 36.09 \\
        w/o KGDT & 34.88 & 29.12 & 36.74 & 45.61 \\
        w/o KGDT (Linear Denoiser) & 36.68 & 31.63 & 39.07 & 47.45 \\
        \bottomrule
    \end{tabular}
    \vspace{-2mm}
\end{table}

\subsection{Ablation Study}
To examine the contribution of each component in M$^2$GDT, we conduct ablation experiments on the MKG-W dataset, as reported in Table~\ref{tab:ablation}. The ablated variants include removing one modality branch at a time (\textit{w/o Text Branch}, \textit{w/o Image Branch}, and \textit{w/o Structure Branch}), removing the semantic anchor branch (\textit{w/o Semantic Anchor}), removing the proposed relation-adaptive semantic routing module (\textit{w/o RASR-MoE}), replacing KGDT with a weaker denoising backbone (\textit{w/o KGDT}), and further replacing KGDT with a simple linear denoiser (\textit{w/o KGDT (Linear Denoiser)}).

Overall, all ablated variants perform worse than the full M$^2$GDT, confirming that the effectiveness of the framework comes from the coordinated interaction among relation-adaptive semantic routing, MLLM-guided alignment, and diffusion-based generation. Removing any modality branch consistently degrades performance, which shows that structural, textual, and visual signals provide complementary cues for multimodal entity completion. Among them, removing the structure branch causes the largest drop (MRR 37.77 \(\rightarrow\) 35.12), indicating that graph structure remains the dominant source of supervision, while textual and visual modalities provide important complementary semantics.

The most severe degradation is observed in \textit{w/o RASR-MoE}, where MRR drops to 30.88. This result verifies the importance of relation-dependent multimodal cue selection, since removing semantic routing substantially weakens the model's ability to select relation-relevant cues before generation. The \textit{w/o Semantic Anchor} variant further reduces MRR to 33.45, showing that anchor-guided semantic calibration is critical for alleviating cross-modal semantic heterogeneity and building a coherent diffusion condition.

The denoising backbone also matters. Replacing KGDT with a weaker denoising backbone reduces MRR to 34.88, while a simple linear denoiser yields 36.68. Although the linear variant remains competitive, both still underperform the full M$^2$GDT, indicating that semantically calibrated conditions alone are not sufficient and that the Transformer backbone is more effective for modeling condition-dependent interactions and complex relational dependencies during iterative denoising.

\subsection{Hyperparameter Sensitivity Analysis}
Figure~\ref{fig:sensitivity} analyzes the sensitivity of M$^2$GDT to four important hyperparameters on MKG-W. Overall, all four curves exhibit clear optima, indicating that the model is stable within a reasonable parameter range rather than being overly sensitive to a single setting.

\noindent\textbf{Diffusion Steps.}
As shown in Figure~\ref{fig:sensitivity}(a), performance improves substantially when the number of diffusion steps increases from 10 to 100, but saturates or slightly declines beyond that point. This suggests that too few steps are insufficient for iterative refinement, whereas excessively many steps bring limited gains while increasing computational cost. So we set the number of diffusion steps to 100.

\noindent\textbf{Transformer Blocks.}
Figure~\ref{fig:sensitivity}(b) shows that increasing the depth of KGDT improves performance up to 3 blocks, after which the gains become marginal. This indicates that a shallow Transformer is insufficient to capture relational dependencies, while deeper architectures provide diminishing returns. We thus use 3 KGDT blocks in the final model.

\noindent\textbf{MLLM Alignment Weight.}
Figure~\ref{fig:sensitivity}(c) shows a clear peak at $\lambda_{2}=0.01$. When the alignment weight is too small, the semantic anchor is not strong enough to effectively calibrate heterogeneous modalities; when it is too large, the alignment objective over-constrains the latent space and reduces the flexibility of diffusion generation. This result supports our view that the MLLM should serve as a semantic anchor rather than a dominant supervisor.

\noindent\textbf{Diffusion Loss Weight.}
As shown in Figure~\ref{fig:sensitivity}(d), the best performance is achieved at $\lambda_{3}=0.1$. A too-small diffusion weight weakens the generative training signal, whereas a too-large value makes denoising dominate the overall objective and suppresses alignment. This confirms that M$^2$GDT requires a balanced interaction between semantic alignment and conditional generation.

\begin{figure}[tbp]
    \centering
    \includegraphics[width=0.50\textwidth]{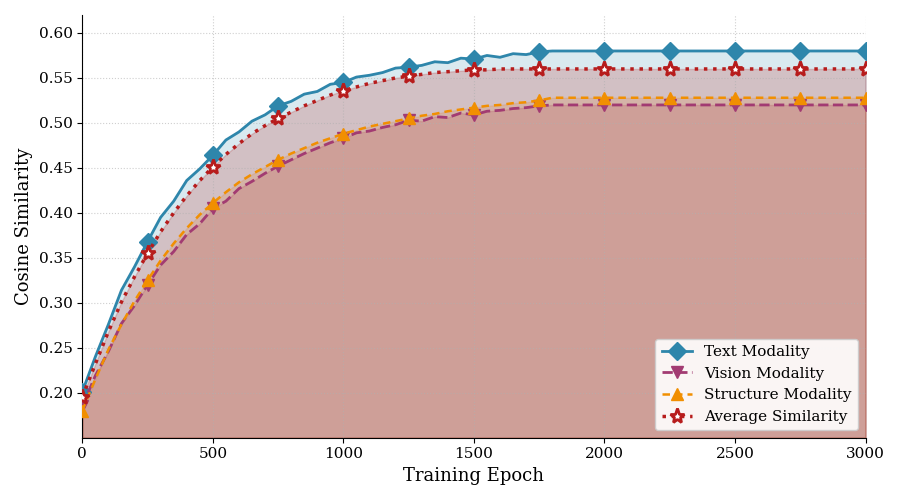}
    % \caption{Evolution of cosine similarity for cross-modal feature alignment during training (0$\sim$3000 epochs) on the MKG-W dataset. 
    % The similarity values rise steadily with slight natural fluctuations and converge stably in 0.52$\sim$0.58 after 1800 epochs. 
    % This moderate alignment verifies a good balance between semantic consistency and structural specificity of knowledge graphs.
    % }
    \caption{Evolution of cosine similarity for text, vision, structure, and average cross-modal alignment during training on the MKG-W dataset.}
    \label{fig:similarity}
\end{figure}
\subsection{Analysis and Generalization}
\begin{table}[tbp]
    \centering
    \small
    \caption{Performance comparison using different MLLMs as semantic anchors on MKG-W dataset.}
    \label{tab:mllm_generalization}
    \begin{tabular}{@{}lcccc@{}}
        \toprule
        \textbf{MLLM Variant} & \textbf{MRR} & \textbf{H@1} & \textbf{H@3} & \textbf{H@10} \\
        \midrule
        \textbf{M$^2$GDT(Qwen3-VL-7B)} & \textbf{37.77} & \textbf{32.18} & \textbf{39.83} & \textbf{48.15} \\
        M$^2$GDT(CLIP-ViT-L/14) & 36.85 & 31.25 & 39.02 & 47.63 \\
        M$^2$GDT(BLIP-2) & 36.42 & 31.89 & 38.67 & 47.21 \\
        \hline
        MCKGC (Previous SOTA) & 36.88 & 31.32 & 38.92 & 47.43 \\
        % MoMoK & 35.89 & 30.38 & 37.54 & 46.13 \\
        \bottomrule
    \end{tabular}
    \vspace{0.2em}
\end{table}

To evaluate the generalization and theoretical rationality of our \textit{align-then-diffuse} paradigm, we analyze its core decoupling design with quantitative similarity metrics and verify compatibility with alternative MLLMs. This paradigm's core advantage is decoupling semantic alignment from structured generation: heterogeneous multimodal features are calibrated into a unified MLLM-guided semantic space, while the diffusion model preserves KG structural specificity, avoiding feature homogenization via moderate cross-modal alignment.

Quantitative validation via cosine similarity (see \figurename~\ref{fig:similarity}) shows that modal similarities rise steadily with minor training fluctuations and converge at 0.52$\sim$0.58 after 1800 epochs, with text ($0.58$) $>$ structure ($0.55$) $>$ vision ($0.52$) and an average of $0.55$. This ranking reflects intrinsic modal differences with MLLM semantics, and the moderate interval balances semantic consistency and structural specificity—avoiding performance loss from over-alignment ($>0.60$) or insufficient alignment ($<0.50$), which underpins our model's generalization ability.

Generalization experiments (Table \ref{tab:mllm_generalization}) show replacing Qwen3-VL-7B with CLIP-ViT-L/14 \cite{radford2021learningtransferablevisualmodels} achieves 36.85\% MRR (0.92\% drop), competitive with SOTA MCKGC (36.88\% MRR). This minimal degradation is consistent with \figurename~\ref{fig:similarity}, where cross-modal similarity fluctuates only by $\pm0.01$ across MLLMs, confirming stable alignment independent of MLLM architecture. Performance gaps across MLLMs derive from their understanding ability: CLIP lacks strong language understanding compared with Qwen3-VL, while even BLIP-2 \cite{li2023blip2bootstrappinglanguageimagepretraining} maintains superiority (36.42\% MRR). 

These results confirm the \textit{align-then-diffuse} paradigm is a robust, transferable foundation for MKGC, adaptable to diverse multimodal backbones for practical deployment. Our core contribution is the decoupling design unifying semantic alignment and structural generation, rather than reliance on a specific MLLM.

\section{Conclusion}
In this paper, we proposed \textbf{ MLLM-Guided Diffusion Transformer with Relation-Adaptive Mixture-of-Experts}, an \textit{align-then-diffuse} framework for \textbf{Multimodal Knowledge Graph Completion}. The framework addresses two key challenges in Multimodal Knowledge Graph Completion: cross-modal semantic heterogeneity and relation-dependent multimodal cue selection. We first introduce a \textbf{Relation-Adaptive Semantic Routing Mixture-of-Experts} module to perform relation-dependent semantic routing over multimodal inputs, then use a frozen \textbf{Multimodal Large Language Model} as a semantic anchor to align the routed representations into a unified latent space, and finally perform graph-conditioned diffusion generation with a \textbf{Knowledge Graph Diffusion Transformer}. In this way, the framework decouples multimodal cue selection, semantic calibration, and structure-aware generation into a coherent pipeline. Extensive experiments validate the effectiveness of the framework.
% Experiments on three benchmark datasets show that the framework consistently outperforms strong baselines, and further analyses verify the effectiveness of the proposed routing, alignment, and diffusion design. 

\bibliographystyle{ACM-Reference-Format}
\bibliography{sample-base}
\end{document}